\documentclass[11pt]{article}
\usepackage{acl2014}
\usepackage{times}
\usepackage{url}
\usepackage{latexsym}

\usepackage{setspace,tikz,linguex,amsmath}
\usepackage{multicol,booktabs,verbatim}

\usepackage{relsize,etoolbox}
\AtBeginEnvironment{quote}{\small}
\AtBeginEnvironment{itemize}{\small}
\AtBeginEnvironment{enumerate}{\small}
\usepackage{enumitem}
\setlist[itemize]{noitemsep}
\setlist[enumerate]{noitemsep}
\setlist[itemize]{label=--}

\makeatletter

\def\citealt{\def\citename##1{{\frenchspacing##1} }\@internalcitec}

\def\@citexc[#1]#2{\if@filesw\immediate\write\@auxout{\string\citation{#2}}\fi
  \def\@citea{}\@citealt{\@for\@citeb:=#2\do
    {\@citea\def\@citea{;\penalty\@m\ }\@ifundefined
       {b@\@citeb}{{\bf ?}\@warning
       {Citation `\@citeb' on page \thepage \space undefined}}%
{\csname b@\@citeb\endcsname}}}{#1}}

\def\@internalcitec{\@ifnextchar [{\@tempswatrue\@citexc}{\@tempswafalse\@citexc[]}}

\def\@citealt#1#2{{#1\if@tempswa, #2\fi}}

\makeatother

\title{\vspace{-\baselineskip}
Detecting and ordering adjectival scalemates\\[0.8em]
\textnormal{\small Paper presented at MAPLEX 2015, February 9-10, Yamagata, Japan (\url{http://lang.cs.tut.ac.jp/maplex2015/})}}

\author{Emiel van Miltenburg\\
  The Network Institute \\
  VU University Amsterdam \\
\texttt{emiel.van.miltenburg@vu.nl}}

\date{}

\begin{document}
\setlength{\Exlabelsep}{0em}
\setlength{\SubExleftmargin}{1em}
\renewcommand{\firstrefdash}{}

\maketitle
\section*{Abstract}
This paper presents a pattern-based method that can be used to infer adjectival scales, such as $\langle\textit{lukewarm,warm, hot}\rangle$, from a corpus. Specifically, the proposed method uses lexical patterns to automatically identify and order pairs of scalemates, followed by a filtering phase in which unrelated pairs are discarded. For the filtering phase, several different similarity measures are implemented and compared. The model presented in this paper is evaluated using the current standard, along with a novel evaluation set, and shown to be at least as good as the current state-of-the-art.

\section{Introduction\footnote{All data from this paper is available online at \url{http://kyoto.let.vu.nl/~miltenburg/public_data/adjectival-scales/}}}
Adjectival scales are sets of (typically gradable) adjectives denoting values of the same property (temperature, quality, difficulty), ordered by their expressive strength \cite{horn1972}. A classical example is $\langle\textit{decent, good, excellent}\rangle.$ In this paper, I also use the term \emph{scale} for ordered sets of non-gradable adjectives, such as $\langle\textit{local, national, global}\rangle$. Scales are ordered such that each adjective is stronger (more informative) than the one preceding it. In this paper, I present a corpus-based method that makes use of lexical patterns to extract pairs of \emph{scalemates}: adjectives that occur on the same scale. As we shall see, due to the nature of the patterns used to extract the scalemates, we also have a reliable way of ordering those pairs. 

What I will not attempt here, is to go beyond scalemates and try to construct full adjectival scales (though see Section\nobreakspace \ref {larger_scales} for some ideas on how to do so). My interest lies in detecting differences in informativeness and expressiveness between adjectives. This is useful e.g.\ for question-answering and information extraction \cite{de2010good}.\footnote{\newcite[808--814]{sheinman2013large} list more applications.} On a more theoretical level, this paper provides the first step in determining which expressions might serve as a stronger alternative to a given adjective. This is useful to diversify the study of scalar inferences (cf. \citealt{doran2009non}). Indeed, this paper finds its origin in the study of scalar diversity \cite{van2014scalar}.

\section{Background}
Now over twenty years ago, \newcite[henceforth H\&M]{hatzivassiloglou1993towards} outlined the first method to semi-automatically identify adjectival scales, producing clusters akin to those in \cite{pantel2003clustering}. Their model consists of the following three steps:

\begin{enumerate}
\item Extract word patterns.
\item Compute word similarity measures.
\item Combine similarities to create clusters of adjectives.
\end{enumerate}

H\&M also suggest to use tests such as Horn's \shortcite{horn1969presuppositional} \emph{X is ADJ, even ADJ} to identify adjectives that are on the same scale (henceforth \emph{scalemates}).\footnote{Similarly, \newcite{hearst1992automatic} later identified hyponyms using lexical patterns.} However, they rejected this idea because ``such tests cannot be used computationally to identify scales in a domain, since the specific sentences do not occur frequently enough in a corpus to produce an adequate description of the adjectival scales in the domain'' (p. 173). In this contribution, I will show that the advent of large corpora made this approach not only feasible, but also competitive with the current state-of-the art.

After H\&K, early work in sentiment analysis attempted to classify documents by determining the average polarity (positivity or negativity) of the words in those documents \cite{turney2002unsupervised}. Research in this direction shows that we can not only obtain clusters of semantically related adjectives (like H\&M do), but we can also determine the semantic orientation of those adjectives. This work stops just short of determining the ordering of scalemates in terms of expressive strength.

\newcite{potts2011developing} provides both a method to categorize words by their orientation, and a method to induce scales. These rely on a data set of online reviews (books, movies, restaurants). The categorization method works as follows. Following the same approach as \newcite{de2010good}, a regression model for the distribution of the ratings is computed for each adjective.\footnote{Potts also studies the polarity of adverbs, but these lie outside the scope of this paper.} Adjectives with a positive correlation with the ratings are categorized as positive, and vice versa. Lacking a significant correlation, adjectives are labeled `neutral.' All words are then ordered by the strength of their coefficients in the regression analysis, after which related adjectives are clustered together using their similar-to's in WordNet \cite{fellbaum1998wordnet}. These clusters are taken to correspond to lexical scales. Potts evaluates his scales on the MPQA subjectivity lexicon \cite{wilson2005recognizing}. In this dataset words are labeled either `strongly subjective' or `weakly subjective.' So for each pair of adjectives $a_1,a_2$, the MPQA lexicon can indicate whether $a_1$ is stronger/weaker than $a_2$ or whether both adjectives have the same score. Comparing his results with the MPQA lexicon, Potts' method achieves a 65\% accuracy on the stronger/weaker items.

Although the results discussed above are very interesting, and certainly deserve further investigation, the focus on sentiment precludes the study of `sentiment-neutral' scales (e.g.\ $\langle\textit{optional, necessary, essential}\rangle$). With our pattern-based method, we provide a more general algorithm that should be able to identify adjectival scales across the board.

\section{A pattern-based approach}\label{patternbased}
Our approach is described in the three sections below. First we describe the basic method, followed by an overview of the measures we implemented to filter the raw data. Finally, we provide a motivation for our choice of corpus.

\subsection{Basic method}
As mentioned in the introduction, we employed a pattern-based method to detect adjectival scales (cf.\ Hearst, 1992). We used the following patterns:

\begin{center}
{\small \begin{tabular}{ll}
-- ADJ$_1$ if not ADJ$_2$ & -- ADJ$_1$ and perhaps ADJ$_2$\\
-- ADJ$_1$ but not ADJ$_2$ & -- between ADJ$_1$ and ADJ$_2$\\
-- from ADJ$_1$ to ADJ$_2$ & -- ADJ$_1$ or at least ADJ$_2$\\
\end{tabular}}
\end{center}

The patterns are tagged with part-of-speech information. These patterns tell us which adjectives are likely to be scalemates, as well as how they are ranked on the scale. In all except the last pattern, ADJ$_1$ is generally weaker than ADJ$_2$, therefore the ordering should be $\langle\textrm{ADJ$_1$, ADJ$_2$}\rangle$. If a pair occurs in two different orders, the most frequent order is kept. On a draw, the pair is discarded.\footnote{There is some room for improvement here. E.g. one could establish a measure of reliability by demanding that the pair is ordered the same way in at least 80\% of the cases. We will not pursue this matter here.}

\subsection{Similarity measures}
The patterns listed above are fairly reliable at identifying scalemates, but no result is perfect. Therefore, we implemented three different types of similarity measures to ensure that the pairs of adjectives are semantically related.

\begin{description}[noitemsep]
\item[LSA \cite{deerwester1990indexing}] If two potential scalemates have a non-negative cosine similarity, they are considered similar.\footnote{We used the TASA model from: \url{http://www.lingexp.uni-tuebingen.de/z2/LSAspaces/}}\singlespacing \vspace{-0.5\baselineskip}

\item[Shared attributes] If two potential scalemates share an attribute, they are considered similar. We used two sets of attributes: \singlespacing
\vspace{-1\baselineskip}
\begin{itemize}[noitemsep]
\item SUMO mappings \cite{pease2002suggested}.
\item WordNet synset attributes. \vspace{-0.1\baselineskip}
\end{itemize}\singlespacing \vspace{-0.5\baselineskip}

\item[Thesaurus] If two scalemates occur in the same thesaurus entry, they are considered similar. We used the following resources: \singlespacing
\vspace{-1\baselineskip}
\begin{itemize}[noitemsep]
\item Lin's \shortcite{lin1998automatic} dependency-based thesaurus.
\item  The Moby thesaurus \cite{ward1996moby}.
\item Roget's thesaurus.\footnote{We used the Jarmasz \& Szpakowicz' \shortcite{jarmasz2001design} \textsc{head} files.}
\end{itemize}
\end{description}  \vspace{-0.5\baselineskip}

\noindent We also implemented two methods to filter the results. These filters are described below.

\begin{description}[noitemsep]
\item[Antonymy] If two potential scalemates are antonyms, they are removed. Antonyms are detected: \singlespacing
\vspace*{-1\baselineskip}
\begin{itemize}[noitemsep]
\item on the basis of their morphology; pairs of the form \{A, prefix-A\} are considered antonyms iff $\textit{prefix} \in \{\textit{il, in, un, im, dis, non-}\}$
\item if they are listed in WordNet as such.
\end{itemize}\singlespacing

\item[Polarity] If two potential scalemates do not share the same polarity in  Hu \& Liu's \shortcite{hu2004mining} opinion lexicon, they are removed.
\end{description}

\subsection{Corpus}
We used the UMBC WebBase corpus \cite[3 bn words]{umbc-corpus} to look up the occurrences of the patterns. The corpus is tagged with part-of-speech data, and its size and scope make it ideal for our purposes.

\section{Results}
We found 32470 pairs of potential scalemates, containing 16971 different adjectives. In general, what we see in the data is that the more patterns a pair occurs in, the more likely it is that the pair consists of two scalemates. Below are some of the pairs that occurred in 5--6 different patterns.

\begin{center}
{\small 
\begin{tabular}{ll}
-- warm hot & -- regional national\\
-- regional global & -- difficult impossible\\
-- weekly monthly & -- unlikely impossible\\
\end{tabular}}
\end{center}

\noindent Compare these with the pairs below, that occurred only in one type of pattern. Some of these pairs are indeed scalemates (e.g.\ \textit{transitive, symmetric}), while others are clearly antonyms (\textit{good, inadequate}).\footnote{One reviewer asks whether the adjectives found in only one pattern are infrequent. While there are pairs containing two rare adjectives, most pairs consist of one frequent and one infrequent adjective, e.g.\ $\langle$\emph{ugly, grotesque}$\rangle$, but there are also examples of pairs with two fairly common adjectives $\langle$\emph{smart, gifted}$\rangle$.}

\begin{center}
{\small 
\begin{tabular}{ll}
-- good inadequate & -- interactive incremental\\
-- affordable scalable & -- damnable devil-ridden\\
-- transitive symmetric & -- ecclesial nonecclesial\\
\end{tabular}}
\end{center}

\noindent As Table\nobreakspace \ref {table:counts} makes clear, most pairs only occur in one type of pattern. What this means is that we cannot do without filtering our results.

\begin{table}[!h]
\begin{center}
\begin{tabular}{lllllll}
\toprule
Patterns & 1 & 2 & 3 & 4 & 5 & 6\\\midrule
Pairs & 29,593 & 2,420 & 336 & 88 & 30 & 3\\
\bottomrule
\end{tabular}
\end{center}
\vspace{-\baselineskip}
\caption{Pairs occurring in $n$ types of patterns.\label{table:counts}}
\end{table}

Table\nobreakspace \ref {table:results} presents the number of pairs that were retrieved for each similarity measure--filtering combination (third column). The table shows big differences in the amount of results between the different similarity measures. Whereas we get 1533 results using Roget's thesaurus, our LSA-based method produces nearly ten times as many pairs of scalemates.

Differences in the amount of results are due to two factors: coverage and lenience. Consider Roget's thesaurus and LSA. \emph{Roget's} is handcrafted, and has a much lower coverage than our automatically generated LSA model. As a consequence, LSA yields a lot more results. Regarding lenience: depending on the similarity measure, the conditions on `being similar' can be more or less lenient. Thesaurus-based measures (Moby, Roget) can be considered strict, demanding near-synonymy. The SUMO measure, on the other hand, is quite lenient; for example, it considers any pair of adjectives that could be considered `subjective assessment attributes' to be similar. Needless to say, with the SUMO measure we get a lot more results. In the case of LSA, we can modify the leniency by raising or lowering the threshold value for the cosine similarity function. We did not experiment with this threshold.

\begin{table}[!h]
\begin{center}
\begin{tabular}{p{1.2cm}p{1.5cm}rrr}
\toprule
Method & Filter & \# Pairs & \# Test & Score\\
\midrule
Raw&None&32,470&2,611&\textbf{60.90}\\
&Antonyms&30,971&2,565&60.55\\
&Polarity&30,628&2,090&59.67\\
&Combined&29,249&2,070&59.42\\
\midrule
Lin&None&8,086&1,027&\textbf{57.84}\\
&Antonyms&7,747&992&57.26\\
&Polarity&7,393&859&56.00\\
&Combined&7,149&844&55.57\\
\midrule
LSA&None&15,233&1,808&\textbf{60.56}\\
&Antonyms&14,682&1,767&60.10\\
&Polarity&14,005&1,463&58.85\\
&Combined&13,561&1,447&58.53\\
\midrule
Moby&None&2,230&287&63.76\\
&Antonyms&2,172&285&\textbf{63.86}\\
&Polarity&2,108&268&62.31\\
&Combined&2,058&267&62.17\\
\midrule
Roget&None&1,533&225&\textbf{62.22}\\
&Antonyms&1,513&224&62.05\\
&Polarity&1,445&203&59.61\\
&Combined&1,430&202&59.41\\
\midrule
SUMO&None&12,061&1,947&\textbf{62.25}\\
&Antonyms&11,498&1,904&61.87\\
&Polarity&10,610&1,548&61.30\\
&Combined&10,152&1,529&61.02\\
\midrule
WordNet&None&1,602&141&\textbf{70.92}\\
&Antonyms&1,384&114&67.54\\
&Polarity&1,402&95&69.47\\
&Combined&1,245&84&66.67\\
\bottomrule
\end{tabular}
\end{center}
\vspace{-\baselineskip}
\caption{Pair counts for each similarity measure, along with MPQA evaluation scores (percentage correct) for each similarity measure--filtering method combination.}
\label{table:results}
\end{table}

\subsection{Evaluation procedure}
\label{literature_eval}
In previous work, evaluation of semantic scales has been done in two ways: intrinsically, using the MPQA lexicon (like Potts 2011), and extrinsically, using the indirect question-answer pairs (IQAP) corpus \cite{de2010good}. An example of an indirect question-answer pair is given in \Next.

\ex. A: Advertisements can be good or bad. Was it a good ad?\\
B: It was a great ad.

To know whether B's answer implies `yes' or `no,' it is necessary to know whether \emph{great} is better than \emph{good} or not.\footnote{\newcite{de2010good} do this in two ways: either using review data, like \newcite{potts2011developing} does as well, or using Web searches. E.g. to answer the question in \Next, De Marneffe et al. searched the Web for `warm weather,' in order to find out the typical range and distribution of degrees associated with warm weather.

\ex. Q: Is it warm outside?\\
A: It's 25$^{\circ}$C

These search results could in theory be compared with those from other queries, allowing for a ranking of temperature-related adjectives. Whether this yields good scales remains to be seen.} In what follows, we will focus on the intrinsic evaluation of our results, as our main goal is to get reliable data. Extrinsic evaluation is left to further research.

Like \newcite{potts2011developing}, we make use of the MPQA sentiment lexicon. For all adjective pairs that contrast in strength according to the lexicon, we check whether our algorithm produces the correct ordering: $\langle weak, strong\rangle$. Because the MPQA lexicon is two-valued, it often occurs that pairs of adjectives have the same label (i.e. are judged equally subjective). This contrasts with Potts' (2011) method, which uses continuous values and thus two adjectives are rarely judged to be equally subjective. As a consequence of this, Potts' model has an overall accuracy of 26\%. We believe that a restriction of the evaluation set to pairs of adjectives that contrast in their subjectivity provides a more reliable assessment of the quality of Potts' data (and thus 65\% accuracy is the score to beat). Either way, the coarse-grainedness of the MPQA lexicon is an issue that needs to be taken into account.

In addition to the MPQA lexicon, we use psychological arousal norms (i.e. values indicating how arousing particular words are), collected by  \newcite[henceforth WKB]{warriner2013norms} for 13,915 English lemmas. The (continuous) arousal values range from 1 (calm) to 9 (aroused). Examples of adjectives with low arousal values are \emph{calm} and \emph{dull}, and \emph{quiet}. Some arousing adjectives are \emph{ecstatic} and \emph{exciting}. Intuitively, the latter have more expressive strength, and as such we can use arousal values as an indication of how scalar expressions should be ordered: $\langle\textit{low, high}\rangle$. Since the WKB data has not been used before in any test of scalarity, we will also compare both evaluation measures to assess their reliability. 

\subsection{Evaluation}\label{sec:evaluation}
Table\nobreakspace \ref {table:results} presents general statistics and the results of the evaluation procedure. The pattern-based method turns out to have a very high recall, with 32,470 different pairs of adjectives. 
Out of all these pairs, 2,611 scalemates have contrasting subjectivity measures in the MPQA database. 1,590 (60.9\%) of these pairs are correctly predicted to be in $\langle\textit{weak, strong}\rangle$ order. A two-tailed Fisher's exact test reveals that the difference between our results and Potts' \shortcite{potts2011developing} data is not statistically significant (p=0.1547).\footnote{Potts achieves 201/308 correct predictions (p.\ 65).}

As presented in rows 2--4 for each method, weeding out antonym pairs and adjectives with opposite polarities does reduce the number of scalemates our algorithm yields, but it does not improve the results. However, this was to be expected: it is not the goal of these filters to improve ordering. %\footnote{It could still be argued that removing all the irrelevant pairs \emph{should} improve the score: by removing the irrelevant pairs, we also remove the noise that might drag performance down. However, our data shows no such improvement.} 
Rather they are meant to exclude pairs of adjectives that are not on the same positive or negative (sub)scale. A different measure is needed to assess the quality of the scales. Likewise, we cannot fully assess which of our different similarity measures is superior. 

The WKB evaluation yields slightly lower scores than those obtained with the MPQA dataset (56--60\%). But how reliable are those scores? To find out, we took the raw scalemates and compared the orderings predicted by the MPQA and WKB datasets. It turns out that they agree on only 62\% of the orderings. This is a surprisingly low number, which casts doubt on the value of these data sets as an individual evaluation metric for adjectival scales. We made the evaluation more robust by combining the two evaluation sets, using only those pairs for which both sets agree on the order. The scores for our algorithm using this new evaluation set is given in Table\nobreakspace \ref {table:eval3}.

\begin{comment}
\begin{table}[!h]
%\vspace{-\baselineskip}
\begin{center}
\begin{tabular}{lrlr}
\toprule
Data & \# Items & Emotion & Score\\
\midrule
WKB & 2079 & arousal & 61.95\\
&& dominance & 57.62\\
&& valence & 55.22\\
\bottomrule
\end{tabular}
\end{center}
\vspace{-\baselineskip}
\caption{Prediction scores (percentage correct) for the three different emotions from the WKB data on the MPQA data set.}\label{table:datadata}
\end{table}
\end{comment}

\begin{table}[!h]
\begin{center}
\begin{tabular}{lrr}
\toprule
Method & \# Items & Score \\
\midrule
Raw&1288&67.49\\
Lin&523&68.50\\
LSA&904&67.32\\
Moby&132&72.73\\
Roget&111&72.07\\
SUMO&1004&68.31\\
WordNet&66&\textbf{77.27}\\
\midrule
Potts 2011 & 74 & 58.11\\
\bottomrule
\end{tabular}
\end{center}\vspace{-\baselineskip}
\caption{Results for the evaluation using only pairs for which the MPQA and WKB data agreed on the ordering. Scores are given for the unfiltered data. Filtering generally had a negative effect on the score of about one percent.}
\label{table:eval3}
\end{table}

The results for our algorithm on this new evaluation set are noticeably (around seven percentage points) better than on either of the datasets alone. How would Potts' \shortcite{potts2011developing} methods score on the improved evaluation set? We expected that his approach might fare better here, as his method relies more on emotion, finding words that express people's feelings about certain products. That seems like an ideal match for an evaluation based on subjectivity and arousal. 
  Our pattern-based method is more general, and also finds (sub-)scales that are not emotion-related (e.g.\ the pair $\langle \textit{\textit{important, crucial}\thinspace}\rangle$). Contrary to our expectations, Potts' method has a lower accuracy, predicting the correct order 58\% of the time (43/74 items).\footnote{\parbox{0.45\textwidth}{Potts' data is available at \url{http://web.stanford.edu/~cgpotts/data/wordnetscales/}.}}

\section{Similar work}
Another pattern-based approach implementing H\&M's ideas is AdjScales, which uses online search engines to determine scale-order  \cite{sheinman2009adjscales,sheinman2013large}. For each pair $\{\textit{head-word, similar-adjective}\}$ in WordNet, Sheinman and colleagues searched the Web using patterns similar to ours to see which ordering was more prevalent. E.g. since \Next[a] returns significantly more results on Google than \Next[b], we may conclude that the ordering should be $\langle \textit{warm, hot} \rangle$. Sheinman et al.\ show that the precision of AdjScales is close to native speaker level.

\vspace{-0.5\baselineskip}

\begin{multicols}{2}
\ex. \a. warm, if not hot
\b. hot, if not warm

\end{multicols}\vspace{-0.5\baselineskip}

The main difference between Sheinman et al.'s work and ours is that Sheinman and colleagues take adjective pairs from WordNet to see how they should be ordered, whereas our method is more agnostic: we use patterns to extract adjective pairs, and only afterwards do we check whether both adjectives are related.\footnote{Theoretically, we can obtain the same results as Sheinman et al.\ by using WordNet's \emph{similar-to} relation as a similarity measure.}
There are three problems with using WordNet as a starting point:

\begin{enumerate}[noitemsep]
\normalsize
\item Not all words are covered by WordNet.\footnote{One reviewer notes that Sheinman et al. do not intend to depart from WordNet, but instead order the adjectives already present in WordNet. With this goal in mind, WordNet's coverage is not an issue. But when the goal is to \emph{enrich} WordNet, or to build a separate lexical resource, we should be able to look beyond WordNet's vocabulary.}
\item Not all related adjectives are related in WordNet, e.g.\ $\langle\textit{difficult, impossible}\rangle$
\item It ignores \emph{ad-hoc} scales \cite{hirschberg1985theory}, made up of words that are not typically related.\footnote{This is relevant for researchers in pragmatics, but of little importance if our goal is to acquire conventional scales. Still, being too restrictive \emph{a priori} may ignore potentially interesting results (cf. problem 2).}%One reviewer notes that, in the latter case, ignoring ad-hoc scales is actually a \emph{benefit} of using WordNet. I agree with this general sentiment, but being too restrictive \emph{a priori} may ignore potentially interesting results (cf. problem 2).}
\end{enumerate}

In our approach, the search space is not constrained by any lexical resource. We simply collect all pairs of adjectives that occur in one of the patterns. To find related pairs that aren't related in WordNet, one can simply choose a different similarity measure. Ad-hoc scales can be found by looking through the raw results, or by choosing a lenient similarity measure.

\section{Future research}
Our results are promising, but as the research on adjectival scales has not received much attention in the literature, there are still many interesting avenues of research. First, there is a clear need for gold standard data, as the available evaluation data is not specifically designed for this task, and show clear shortcomings (e.g.\ coarse-grainedness ---as discussed in Section\nobreakspace \ref {literature_eval}, which precludes the evaluation of scalemates that are very close in terms of expressive strength.) Second, there is the possibility of extending our work to other languages. Third, I see a lot of potential in using vector-based approaches to generate ordered scales from a corpus. I discuss these issues in turn. Finally, I consider the possibility of constructing larger scales from our set of scalemates.

\subsection{Creating a gold standard}
We need to have a real gold standard containing pairs of scalemates annotated with their ordering and polarity. This gold standard should be balanced in terms of emotion-related scales and other kinds of scales. We believe that the data generated using our pattern-based method, combined with Potts' \shortcite{potts2011developing} data should provide a good starting point for building a reliable lexical resource.

After we finished our data-analysis, Christopher Potts (p.c.)\ shared the results of an online experiment carried out on Amazon's Mechanical Turk. Participants were shown a set of adjective pairs, and asked for each pair to judge whether the first adjective is stronger, weaker or as strong as the second adjective. This is exactly the kind of data we need to evaluate the order of automatically identified scales. Table\nobreakspace \ref {pottsdata} shows the agreement between our proposed evaluation set (combining the MPQA and WKB data) and the elicited data, followed by the results of  our algorithm on Potts' data. We observe that the combination of the MPQA and WKB data provides a reasonable estimate of the correct ordering of adjectival scales, but our algorithm does much better on the elicited data than on our proposed evaluation set.

\begin{table}[h!]
\begin{center}
\begin{tabular}{lrrrrrr}
\toprule
\multicolumn{6}{c}{Our proposed data set}\\
\midrule
Agreement &   6    &  7    &  8    &  9    &  10 \\
 \# Test items       & 63    & 49    & 36    & 28    &  16 \\
 Accuracy           & 84 & 88 & 92 & 93 & 100 \\
\midrule
 \multicolumn{6}{c}{Pattern-based search}\\
\midrule
 Agreement    &  6   &  7    &  8    &  9   & 10    \\
 \# Test items      & 40   & 36    & 28    & 23   & 15    \\
 Accuracy           & 78 & 83 & 89 & 91 & 93 \\
\bottomrule
\end{tabular}
\end{center}
\caption{Results for our new evaluation set and our algorithm on the evaluation data provided by Christopher Potts (p.c.). The columns correspond to the level of agreement between participants. I.e. how many participants (out of 10) agreed on the first adjective being either stronger or weaker than the second. Making this requirement more strict reduces the amount of test items that we could use, but increases the precision of our evaluation set and our algorithm.}
\label{pottsdata}
\end{table}

There are still two problems with Potts' data set: (i) it is limited in size, and (ii) it is based on Potts'~\shortcite{potts2011developing} study on reviews, and as such is limited in coverage (i.e. it has no `sentiment-neutral' scales). We are planning to expand the set of gold standard data in the future.

\subsection{Other languages \& automation}
In this paper, we have only looked at English scales. How would one go about extracting scales in other languages? Could we further automate our algorithm to generate scales for multiple languages at once? This requires a way to automatically detect patterns in which pairs of scalemates are likely to occur. There are two ways of doing so, both using sets of known scalemates: (i) \newcite{sheinman2009adjscales} take 10 seed word pairs, and extract only those patterns that fulfill certain conditions (e.g. appearing with at least 3 different seed pairs, occurring more than once for each pair, not being restricted to one meaning domain). \newcite{schulam2010automatically} show how this approach can be applied to German.
(ii) \newcite{lobanova2012anatomy} takes a probabilistic approach, estimating the likelihood of patterns to contain one of the seed pairs. She applies this method to find patterns likely to contain antonyms, but her approach can easily be extended to the scale-domain. It may also be fruitful to try a hybrid approach, combining the two. 

Once we have scale ordering data for multiple languages, it should be possible to automatically verify the results through EuroWordNet \cite{vossen2004eurowordnet}, using the Interlingual Index (ILI): intuitively, corresponding synsets should have the same ordering relation in all languages.

\subsection{Semantic vectors}
\newcite{mohtarami2012sense} create a semantic vector space with twelve basic emotions as its dimensions. The position of each word $w_{n}$ in this space is determined by the co-occurrence counts of $w_{n}$ with words in the synsets of the selected basic emotion words. The authors use this information to compute what they call `word pair sentiment similarity.' On the basis of this similarity measure, words expressing similar emotions can be clustered together. While the authors do not go into this, the right ordering of a set of adjectives might be achieved by maximizing the sentiment similarity between all neighboring pairs of adjectives within a cluster. \newcite{kimderiving} provide a more general vector-based method to order adjectives on a scale. Making use of earlier observed semantic regularities in neural embeddings \cite{mikolov2013linguistic}, the authors show how a scale can be generated by extracting words that are located at intermediate points between two vectors from antonym pairs. Though the results (using the IQAP corpus) look promising, the extraction of scalemates has not yet been done on a larger scale.

\subsection{Building larger scales}\label{larger_scales}
A naive way to build scales from pairs of scalemates would be to chain them together, so that e.g. $\langle\textit{lukewarm, warm}\rangle$ and $\langle\textit{warm, hot}\rangle$ could be used to form $\langle\textit{lukewarm, warm, hot}\rangle$. But this strategy completely ignores polysemy. Consider the pairs $\langle\textit{inexpensive, cheap}\rangle$ and $\langle\textit{cheap, rubbish}\rangle$. Together, these yield the incoherent scale $\langle\textit{inexpensive, cheap, rubbish}\rangle$ that mixes up two dimensions: \textsc{cost} and \textsc{quality}. A solution to this issue might be to only chain scales if the senses of the adjectives involved are in the same domain (either verified through WordNet, or using an automatic sense clustering algorithm such as CBC \cite{pantel2003clustering}). However, after crossing that hurdle we run into the problem that scales are highly context-dependent. It might be best to construct adjectival scales \emph{on the fly}: rather than having a stored list of full-blown scales, build a scale consisting of adjectives that are relevant to the discourse. A minimal requirement for such a process is to have pairwise ordering information for all adjectives involved, which is what our pattern-based method produces. 

\section{Conclusion}
In this paper we have looked at different methods to automatically find scales or scalemates from a corpus since H\&K's original paper. Our findings show that a pattern-based method can be very successful at identifying pairs of scalemates, as long as the corpus is big enough. This mirrors findings from \newcite{sheinman2009adjscales} and \newcite{sheinman2013large}. One of our contributions is the use of a wide range of similarity measures as well as an antonymy filter and a polarity filter to clean up the results.

We have also proposed a new evaluation method, combining the MPQA subjectivity lexicon with the WKB arousal norms. The combination of these two data sets makes the evaluation of scale ordering methods more reliable. This alleviates, but does not eliminate the need for a true gold standard, which could finally enable us to move towards the automatic identification of adjectival scales. 

\section*{Acknowledgments}
Thanks to members of the CLTL-group at the VU University Amsterdam for discussion, to Chris Potts for sharing his data, and to Bob van Tiel and two anonymous reviewers for their comments. This research was carried out in the \emph{Understanding Language by Machines} project, made possible through the NWO Spinoza prize awarded to Piek~Vossen.

\bibliographystyle{acl}
\bibliography{textpatterns}

\end{document}